\documentclass[runningheads]{llncs}
\usepackage[review,year=2026,ID=9001]{eccv}
\usepackage{eccvabbrv}
\usepackage{graphicx}
\usepackage{booktabs}
\usepackage[accsupp]{axessibility}
\usepackage{amsmath}
\usepackage{amssymb}
\usepackage{mathtools}

\usepackage{colortbl}
\usepackage{multirow}
\usepackage{microtype}
\usepackage{tabularx}
\usepackage{ragged2e}
\usepackage{makecell}
\usepackage{subcaption}
\usepackage{placeins}
\usepackage[pagebackref,breaklinks,colorlinks,citecolor=eccvblue]{hyperref}
\usepackage[capitalize,noabbrev]{cleveref}
\makeatletter
\renewcommand\section{\@startsection{section}{1}{\z@}%
  {-8pt plus -2pt minus -2pt}{4pt plus 2pt minus 2pt}%
  {\normalfont\large\bfseries}}
\renewcommand\subsection{\@startsection{subsection}{2}{\z@}%
  {-6pt plus -2pt minus -2pt}{3pt plus 1pt minus 1pt}%
  {\normalfont\normalsize\bfseries}}
\renewcommand\paragraph{\@startsection{paragraph}{4}{\z@}%
  {4pt plus 1pt minus 1pt}{-1em}%
  {\normalfont\normalsize\bfseries}}
\makeatother
\begin{document}
\title{Detecting AI-Generated Images via Contextual Anomaly Estimation in Masked AutoEncoders}
\titlerunning{CINEMAE}
\author{Minsuk Jang\inst{1} \and
Firstname2 Lastname2\inst{1,2} \and
Firstname3 Lastname3\inst{2} \and
Firstname4 Lastname4\inst{3} \and
Firstname5 Lastname5\inst{1} \and
Firstname6 Lastname6\inst{1,2,3} \and
Firstname7 Lastname7\inst{2} \and
Firstname8 Lastname8\inst{3} \and
Firstname9 Lastname9\inst{1,2}}
\authorrunning{F.~Lastname1 et al.}
\institute{Department of XXX, University of YYY, Location, Country \and
Company Name, Location, Country \and
School of ZZZ, Institute of WWW, Location, Country}
\maketitle\begin{abstract}
Context-based detection methods such as DetectGPT achieve strong generalization in identifying AI-generated text by evaluating content compatibility with a model's learned distribution. In contrast, existing image detectors rely on discriminative features from pretrained backbones such as CLIP, which implicitly capture generator-specific artifacts. However, as modern generative models rapidly advance in visual fidelity, the artifacts these detectors depend on are becoming increasingly subtle or absent, undermining their reliability. Masked AutoEncoders (MAE) are inherently trained to reconstruct masked patches from visible context, naturally modeling patch-level contextual plausibility akin to conditional probability estimation, while also serving as a powerful semantic feature extractor through its encoder. We propose CINEMAE, a novel architecture that exploits both capabilities of MAE for AI-generated image detection: we derive per-patch anomaly signals from the reconstruction mechanism and extract global semantic features from the encoder, fusing both context-based and feature-based cues for robust detection. CINEMAE achieves highly competitive mean accuracies of 96.63\% on GenImage and 93.96\% on AIGCDetectBenchmark, maintaining over 93\% accuracy even under JPEG compression at QF=50.
\end{abstract}
\section{Introduction}
\label{sec:intro}
AI-Generated Content (AIGC), referring to fully-generated images distinct from partially manipulated forgeries~\cite{du2025forensichub}, has become a pressing societal concern as modern generative models such as diffusion models~\cite{Rombach_2022_CVPR,ho2020denoising, song2021denoising} produce remarkably realistic synthetic data, raising concerns about misinformation, copyright infringement, and ethical misuse~\cite{lyu2024deepfake}.
Consequently, developing reliable detectors for AI-generated images has become a critical research priority.

Supervised detectors~\cite{wang2020cnn, zhang2019detecting} have dominated AIGC detection but tend to overfit to specific generation families (\eg, GANs), limiting generalization to unseen paradigms~\cite{corvi2023detection, yan2025chameleon}.
Recent approaches leverage pre-trained vision backbones (\eg, CLIP~\cite{radford2021learning,ojha2023towards}) and lightweight image transformations~\cite{li2025safe} for cross-generator generalization. 
While effective, these methods still depend on artifact-level cues implicitly captured by the backbone's feature space. However, as modern generative models rapidly advance in visual fidelity, such artifacts are becoming increasingly subtle or absent, calling for detection strategies grounded in contextual plausibility rather than residual traces.
\begin{figure}[t]
    \centering
    \includegraphics[width=0.65\linewidth]{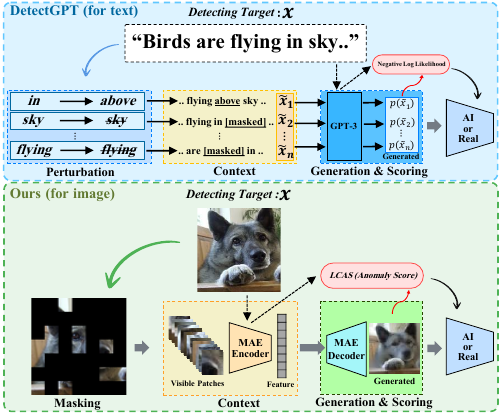}
    \caption{\textbf{Comparison Overview.} Both methods score 
    generated outputs conditioned on context. DetectGPT 
    evaluates perturbed text probability given surrounding words; 
    CINEMAE evaluates reconstructed patch plausibility given 
    visible patches.}
    \label{fig:fig1}
\end{figure}
  
In parallel, the paradigm of context-based detection, pioneered in the text domain with methods like DetectGPT~\cite{mitchell2023detectgpt}, offers a powerful alternative by measuring content's compatibility with a model's learned distribution. 
However, translating this to the visual domain is challenging. ZED~\cite{zed2024eccv} treats average-pooled patches as context to compute distributional conformity, but this pooling-based approach has limited capacity to capture fine-grained patch-level semantics. 
Conversely, prior works using Masked AutoEncoders (MAE)~\cite{he2022masked} have treated the model as a generic feature extractor for downstream classification~\cite{das2023unmasking}, focusing on encoder representations rather than exploiting reconstruction statistics directly. 
Building on this observation, we draw an analogy to DetectGPT, which treats surrounding words as context to score generated text. Similarly, MAE's reconstruction mechanism naturally models contextual plausibility: visible patches serve as context, and the reconstruction discrepancy serves as an anomaly signal indicating whether the masked content is consistent with natural image statistics.
We propose \textbf{CINEMAE}, \textbf{C}ontextual \textbf{I}mage a\textbf{N}omaly \textbf{E}stimation with \textbf{M}asked \textbf{A}uto\textbf{E}ncoder. 
To the best of our knowledge, this is the first work to directly use MAE's reconstruction discrepancy as a per-patch contextual anomaly signal for AIGC detection. Our main contributions are as follows:
\begin{itemize}
    \item We demonstrate the feasibility of context-based detection in the image domain by showing that MAE's reconstruction discrepancy serves as an effective per-patch anomaly signal for AIGC detection, providing a detection basis that is inherently independent of generator-specific artifacts.
    \item We propose \textbf{CINEMAE}, an architecture that exploits MAE's dual capability: per-patch contextual anomaly scoring from the decoder and global semantic feature extraction from the encoder, combining both for robust detection without task-specific fine-tuning.
    \item CINEMAE achieves strong cross-generator generalization, maintaining over \textbf{95.9\%} accuracy on all 8 GenImage generators and over \textbf{91\%} on 15 of 16 unseen generators in AIGCDetectBenchmark, while preserving robustness under JPEG compression at QF=50.
\end{itemize}

\section{Related Works}
\label{sec:related}
\subsection{From Image Forensics to AIGC Detection}
Image forensics has progressed through distinct stages: from detecting localized manipulations such as copy-move and splicing~\cite{zhou2018learning, kwon2021cat}, to Deepfake detection targeting facial identity replacement or reenactment~\cite{afchar2018mesonet, rossler2019faceforensics++, du2025forensichub}. 
With the rapid advance of modern generative models, AI-Generated Content (AIGC) detection has emerged as a new frontier, where the entire image is synthesized from scratch rather than partially manipulated~\cite{du2025forensichub}. This distinction is critical: unlike localized forgeries or face-swapping Deepfakes, AIGC detection requires modeling holistic image-level authenticity signals.
\subsection{Artifact-Driven AIGC Detection}
Most AIGC detectors rely on discriminative features that implicitly capture generator-specific artifacts. Early supervised approaches~\cite{wang2020cnn, zhang2019detecting} overfit to specific generation families, while recent methods leverage CLIP's pretrained feature space for cross-generator generalization: UnivFD~\cite{ojha2023towards} uses nearest neighbor classification in CLIP space, AIDE~\cite{yan2025chameleon} integrates low-level pixel statistics with CLIP features, and C2P-CLIP~\cite{tan2025c2p} injects prompts at training time. Reconstruction-based methods like DRCT~\cite{drct_icml24} and LaRE~\cite{Luo2024LaRE2} exploit latent space reconstruction errors as detection signals.
Despite this progress, these approaches fundamentally depend on artifact cues or frozen vision-language representations. 
\subsection{Context-Based Detection Methods}
In the text domain, context-based detection is pioneered by DetectGPT~\cite{mitchell2023detectgpt}, which identifies AI-generated content by evaluating how a text's log-probability changes under small perturbations. 
ZED~\cite{zed2024eccv} adapted this paradigm to images by computing conditional Negative Log-Likelihood (NLL) from average-pooled patches across a multi-resolution hierarchy to assess distributional conformity. 
While effective at capturing global distributional signals, this pooling-based approach operates across resolution levels, which limits its capacity to capture fine-grained patch-level semantic consistency at the original resolution. 
Separately, recent works have adapted Masked AutoEncoders (MAE)~\cite{he2022masked} for visual forensic tasks: MARLIN~\cite{Cai_2023_CVPR} learns facial video representations via masked reconstruction, while MIFAE-Forensics~\cite{Wang_2025_ICASSP} combines masked image and frequency autoencoding for deepfake detection. 
However, these approaches employ MAE as a static feature extractor, leaving the decoder's reconstruction statistics unexploited as a detection signal.
In contrast, our method directly exploits MAE's reconstruction discrepancy to model conditional relationships between visible and masked patches at full resolution.

\begin{figure}[t]
  \centering
  \begin{subfigure}{0.24\linewidth}
    \centering\includegraphics[width=\linewidth]{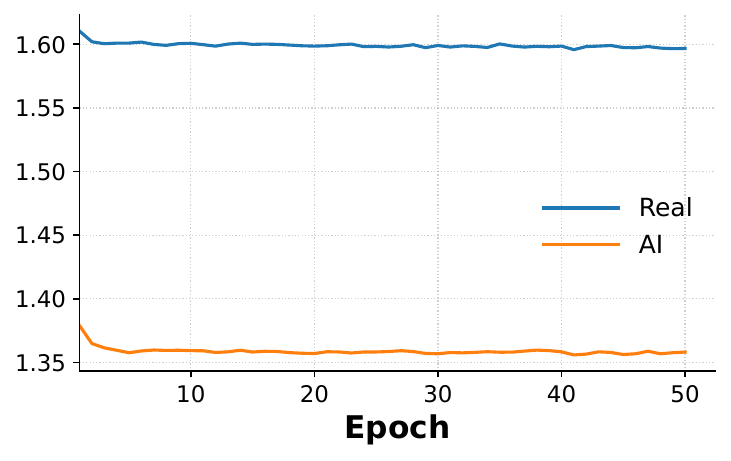}
    \caption{BigGAN}
  \end{subfigure}\hfill
  \begin{subfigure}{0.24\linewidth}
    \centering\includegraphics[width=\linewidth]{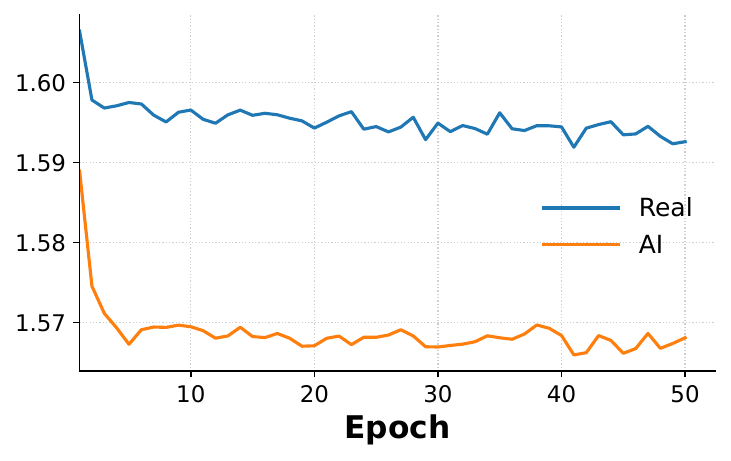}
    \caption{Wukong}
  \end{subfigure}\hfill
  \begin{subfigure}{0.24\linewidth}
    \centering\includegraphics[width=\linewidth]{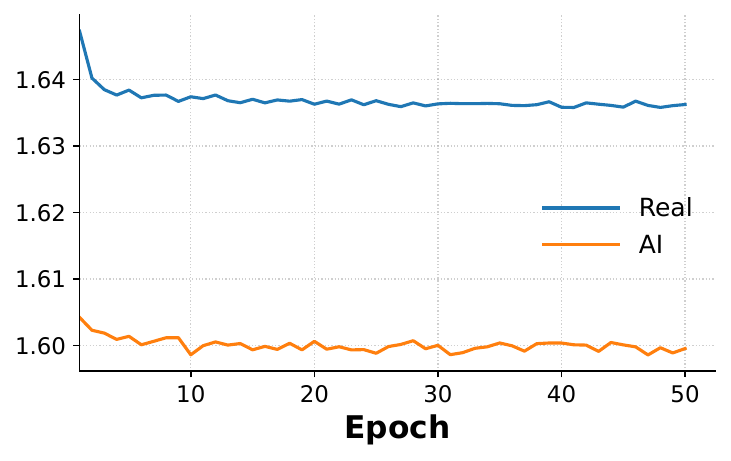}
    \caption{Midjourney}
  \end{subfigure}\hfill
  \begin{subfigure}{0.24\linewidth}
    \centering\includegraphics[width=\linewidth]{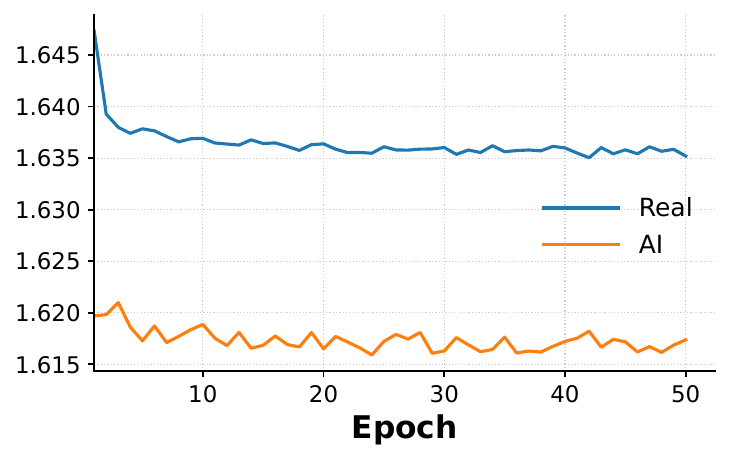}
    \caption{SD v1.4}
  \end{subfigure}
    \caption{Mean patch-level $\text{NLL}_{\text{approx}}$ during decoder training (MSE loss, unfrozen), showing the gap between predictable AI images and complex real images.}
  \label{fig:nll_epoch}
\end{figure}

\section{Motivation}
\label{sec:motivation}
\paragraph{MAE as a Contextual Plausibility Estimator.}
We argue that MAE's reconstruction mechanism can serve as a quantitative measure of contextual plausibility for AIGC detection. MAE reconstructs masked patches conditioned solely on visible patches, meaning the reconstruction error directly reflects how well masked content conforms to the surrounding context under the model's learned distribution of natural images. 
We show below that this reconstruction error is mathematically equivalent to an approximated negative log-likelihood, providing a principled basis for context-based detection in the image domain.
\paragraph{Implicit Likelihood in MAE.}
Let $f_\theta$ denote the MAE decoder. MAE is trained with Mean Squared Error (MSE) loss, which is mathematically equivalent to maximizing Gaussian log-likelihood under fixed variance:
\begin{equation}
    \mathcal{L}_{\text{MSE}} = \|x_m - \hat{x}_m\|^2 \propto -\log p(x_m | x_v; f_\theta) + \text{const},
    \label{eq:mse_gaussian}
\end{equation}
where $x_m$ denotes a masked patch, $x_v$ are visible patches (context), $f_\theta$ is the MAE decoder, and $p(x_m|x_v; f_\theta) = \mathcal{N}(\hat{x}_m, \sigma^2 I)$ is the implicit Gaussian distribution. Thus, MAE inherently models reconstruction errors as Gaussian-distributed during training. Since MAE reconstructs masked patches $\hat{x}_m$ conditioned on visible patches $x_v$, the reconstruction error naturally approximates a context-conditional likelihood. We formalize this as an Approximated Negative Log-Likelihood ($\text{NLL}_{\text{approx}}$):
\begin{equation}
    \text{NLL}_{\text{approx}}(x_m) = \frac{\|x_m - f_\theta(x_v)\|^2_2}{2\sigma^2}\approx-\log p(x_m | x_v; f_\theta) .
    \label{eq:nll_approx}
\end{equation}
 
Crucially, MAE's masking mechanism is inherently conditional by design: it predicts masked patches given visible context. This makes MAE \textbf{a latent conditional likelihood operator}, naturally suited for context-based anomaly detection. 
Following DetectGPT~\cite{mitchell2023detectgpt}, we interpret this not as a calibrated probability but as a compatibility score measuring how well a patch fits its surrounding context under MAE's learned distribution. 
Under this interpretation, real images should yield high $\text{NLL}_{\text{approx}}$ due to their complex local statistics~\cite{Ruderman1994}, while AI-generated images should yield lower scores due to their more predictable structures.

\begin{figure}[t]
    \centering
    \begin{minipage}[t]{0.54\linewidth}
        \centering
        \includegraphics[width=\linewidth, trim={0pt 10pt 0pt 0pt}, clip]{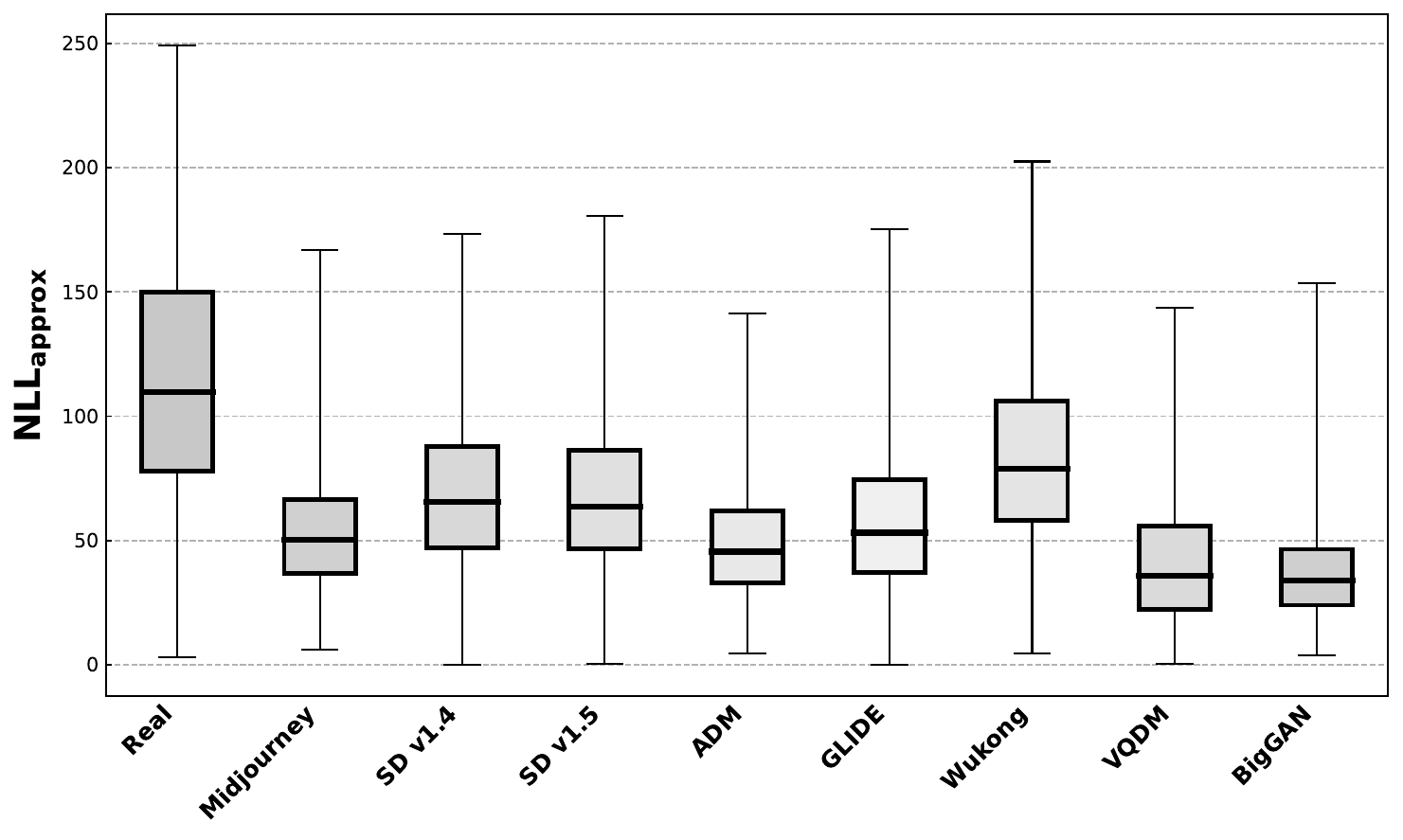}
        \caption{\textbf{Image-level $\text{NLL}_{\text{approx}}$ distributions} from a single forward pass with a fully frozen MAE. Despite different absolute scales from Fig.~\ref{fig:nll_epoch}, the discriminative gap persists.}
        \label{fig:nll_distri}
    \end{minipage}
    \hfill
    \begin{minipage}[t]{0.42\linewidth}
        \centering
        \includegraphics[width=\linewidth]{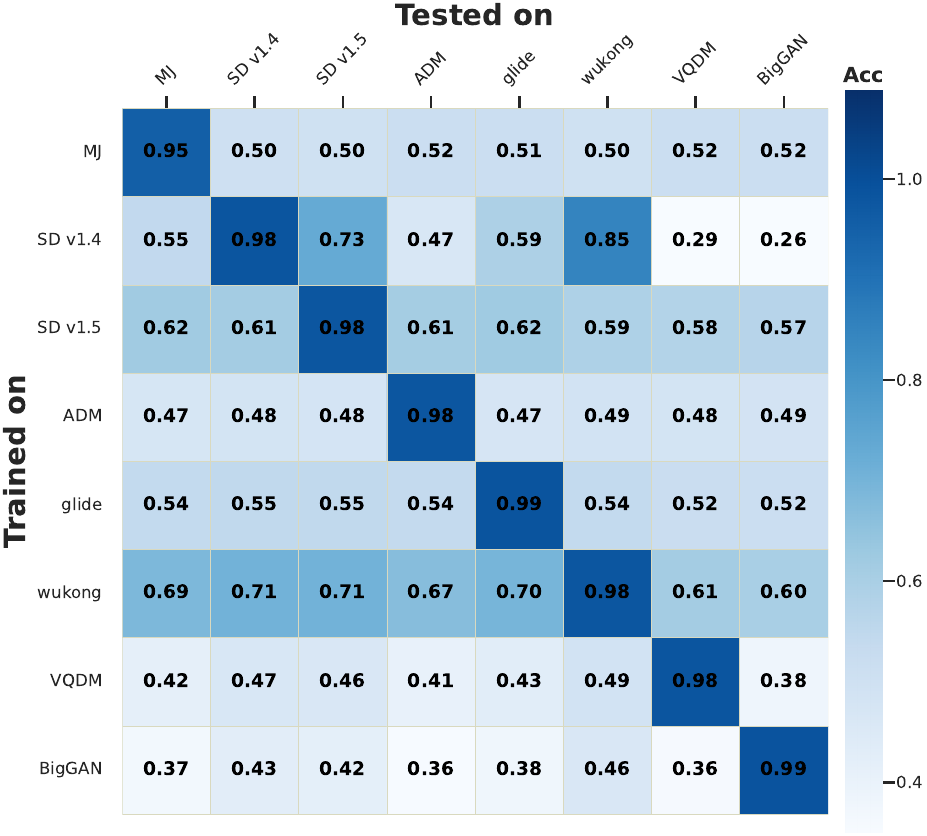}
        \caption{\textbf{Cross-domain generalization breakdown.} Each row: classifier trained on one generator, evaluated on all in GenImage.}
        \label{fig:cross_domain_heatmap}
    \end{minipage}
\end{figure}

\paragraph{Empirical Validation.}
We validate this by monitoring patch-level $\text{NLL}_{\text{approx}}$ during MAE decoder training (ViT-L/16, ImageNet-pretrained encoder frozen, decoder unfrozen to verify that the discriminative gap is intrinsic rather than an artifact of optimization).

As shown in Fig.~\ref{fig:nll_epoch}, across diverse generators (BigGAN~\cite{brock2018large}, Midjourney~\cite{midjourney2022}, Stable Diffusion v1.4 (SD v1.4)~\cite{Rombach_2022_CVPR}, Wukong~\cite{wukong2022}), AI-generated images consistently yield lower $\text{NLL}_{\text{approx}}$ than real images. As training progresses, $\text{NLL}_{\text{approx}}$ converges yet a clear gap persists, with real images remaining at a higher bound due to their inherently complex local statistics. This gap confirms that $\text{NLL}_{\text{approx}}$ captures a discriminative signal for AIGC detection.

Importantly, this gap is not an artifact of decoder training: Fig.~\ref{fig:nll_distri} shows that even a single forward pass through a fully frozen MAE produces clearly separated $\text{NLL}_{\text{approx}}$ distributions between real and AI-generated images, confirming that the discriminative signal is intrinsic to MAE's pretrained natural image prior.

\paragraph{Limitations of $\text{NLL}_{\text{approx}}$ Only Detection.}
However, different generators produce distinct $\text{NLL}_{\text{approx}}$ distributions (Fig.~\ref{fig:nll_distri}). 
To investigate whether a single threshold suffices, we train threshold classifiers on each generator in GenImage~\cite{zhu2024genimage} and evaluate across others (Fig.~\ref{fig:cross_domain_heatmap}): in-domain accuracy reaches 99\%, but cross-domain drops to 26\% (\eg, SD v1.4 $\rightarrow$ BigGAN). 

These analyses reveal that while $\text{NLL}_{\text{approx}}$ captures discriminative signals, it alone cannot generalize across generators due to distribution shifts. 
We therefore propose a two-pathway architecture that fuses patch-level anomaly scores with global MAE features to achieve cross-generator generalization.

\section{Method}
\label{sec:method}
\begin{figure*}[t]
\centering
\includegraphics[width=1.2\textwidth, trim=50 0 10 0, clip]{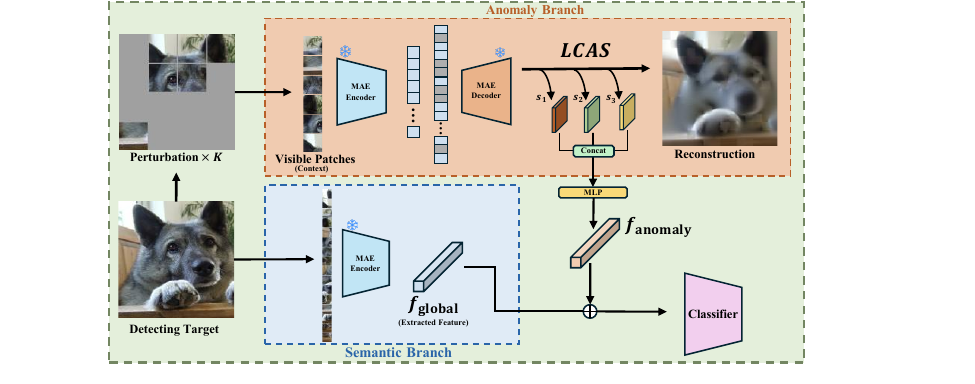}
\caption{Overview of our method. The MAE encoder extracts global features. The decoder reconstructs masked patches to compute patch-wise Local Contextual Anomaly Score (\textit{LCAS}), which are aggregated into anomaly statistics $s_1, s_2, s_3$, fused with global features, and passed to a classifier for binary prediction.}
\label{fig:architecture}
\end{figure*}

We now detail \textbf{CINEMAE}, built on an ImageNet~\cite{deng2009imagenet} pre-trained 
MAE ViT-L/16~\cite{he2022masked} backbone.
We treat this backbone as a generic \emph{natural image prior} that encodes contextual regularities 
of natural images. \textbf{CINEMAE} combines a global semantic 
representation with a local anomaly signal to support cross-generator generalization, where a lightweight fusion 
module uses the anomaly signal as a corrective term to the global representation, as illustrated in Fig.~\ref{fig:architecture}.
\subsection{Architectural Components}
\paragraph{Semantic Branch (Global Representation).}
To capture the overall semantic content of the image, this branch leverages the Vision Transformer's intrinsic mechanism for global feature aggregation. Given an input image $I$, the MAE encoder processes $I$ and produces a sequence of patch-level features. Following standard ViT practice, we extract the global representation from the [CLS] token at the final encoder layer:
\begin{equation}
f_{\text{global}} = \text{Encoder}_{\text{MAE}}(I)_{\text{[CLS]}} \in \mathbb{R}^{D}.
\label{eq:f_global}
\end{equation}
This vector effectively summarizes high-level semantic information from all patches. 
\paragraph{Anomaly Branch (Local Signal).}
This branch quantifies fine-grained contextual anomalies by analyzing the image's patch-level reconstruction statistics using the MAE framework. This process yields a set of local anomaly scores, which are aggregated into a final image-level anomaly feature vector $f_{\text{anomaly}}$ (formally defined in \cref{sec:scores_to_features}).
\subsection{Local Contextual Anomaly Score}
Building on $\text{NLL}_{\text{approx}}$ (\cref{sec:motivation}), which captures discriminative signals but lacks cross-generator generalization, we formulate a more robust anomaly score. To quantify how ``out-of-place'' a given patch is, we formulate our \textbf{Local Contextual Anomaly Score (\textit{LCAS})} as a combination of two distinct but complementary terms: a statistical deviation term and a reconstruction fidelity term. This hybrid score assesses a patch's plausibility by measuring its consistency with both its immediate local surroundings and the image's broader semantic context.

Formally, for a masked original patch $p_i^m$, where $i$ denotes the patch index, we define its anomaly with respect to the set of visible patches $P_v$ (the context) as:
\begin{equation}
\textit{LCAS}(p_i^m, P_v) =
\underbrace{\mathcal{D}_{\text{stat}}(\hat{p}_i^m, P_v)}_{\text{Statistical Deviation}} +
\underbrace{\lambda \cdot \| \hat{p}_i^m - p_i^m \|_2^2}_{\text{Reconstruction Fidelity}},
\label{eq:cas}
\end{equation}
where $\hat{p}_i^m$ is the patch reconstructed by the MAE decoder, and $\lambda$ is a weighting hyperparameter.
The first term, $\mathcal{D}_{\text{stat}}$, measures the statistical deviation of a patch from its visible context, adapting the negative log-likelihood framework from text detection~\cite{mitchell2023detectgpt} to images. This formulation is consistent with MAE's MSE training objective, which implicitly models reconstruction errors as Gaussian-distributed. 
Crucially, the reconstructed patch $\hat{p}_i^m$ is conditioned on visible context $P_v$, not the original patch. 
\textbf{This follows DetectGPT's context-based scoring principle}, but extends it to the image domain by additionally incorporating a feature-level reconstruction fidelity term. 
This score quantifies how ``unexpected'' the reconstruction is, given the per-pixel mean and variance $\{\mu_v, \sigma_v\}$ computed from $P_v$:
\begin{equation}
\mathcal{D}_{\text{stat}}(\hat{p}_i^m, P_v) = 
\frac{1}{2} \left( 
\left( \frac{\hat{p}_i^m - \mu_v}{\sigma_v} \right)^2 
+ \log \sigma_v^2 
\right).
\label{eq:d_stat}
\end{equation}

Here, $P_v$ comprises all visible patches in the image.  For each visible patch, we compute its internal mean and  standard deviation; $\{\mu_v, \sigma_v\}$ are then obtained  by averaging these statistics across all visible patches,  yielding scalar reference values that characterize the  typical patch-level complexity of the image.

The second term measures \textbf{reconstruction fidelity}. 
By penalizing the $\ell_2$ distance between the original masked patch and its MAE-reconstructed counterpart, this term assesses semantic consistency. 
A large reconstruction error indicates that the patch violates the global, high-level context of the image as understood by the MAE, signaling a potential semantic anomaly. 
Collectively, our \textit{LCAS} leverages this duality: an unusually low statistical deviation score suggests unnatural local predictability, while a high reconstruction error points to a global semantic violation. 
This two-pronged approach allows CINEMAE to detect a wider range of artifacts.
\subsection{From Anomaly Scores to Image-Level Features}
\label{sec:scores_to_features}
While the patch-level score \textit{LCAS} (Eq.~\ref{eq:cas}) is effective at identifying local inconsistencies, a single high-scoring patch is not sufficient to classify an entire image. 
An image-level representation must also consider the distribution, magnitude, and stability of these anomalies across the image. 
Therefore, to form a robust, fixed-size feature vector, we aggregate the \textit{LCAS} from all masked patches into a compact anomaly statistics vector. 
Furthermore, inspired by perturbation-based methods in uncertainty estimation, we repeat this process $K$ times with independent perturbation to assess the stability of the anomaly signal. 

This yields three complementary statistics $s_1, s_2, s_3$, where each component is derived from the set of patch-level \textit{LCAS} and captures a different aspect of the image's anomalous nature. Let $\ell_i^{(k)}$ denote the \textit{LCAS} value of the $i$-th patch under the $k$-th random mask ($k \in \{1, \ldots, K\}$), and let $M_k$ be the set of masked patch indices for run $k$:
\begin{itemize}
    \item \textbf{$s_1$ (Variability):} The difference between the maximum and minimum \textit{LCAS} value $\{\ell_i^{(k)}\}_{i \in M_k}$ observed across the masked patches, averaged over the $K$ runs. This metric reflects the spatial dispersion of anomalies, distinguishing between concentrated, high-intensity artifacts and more diffuse inconsistencies.
    \item \textbf{$s_2$ (Overall Anomaly):} The mean of the \textit{LCAS} values across all masked patches and all $K$ runs, \ie, the global average of $\ell_i^{(k)}$ over $k$ and $i \in M_k$. This provides a global measure of the image's total contextual and semantic implausibility.
    \item \textbf{$s_3$ (Perturbation Sensitivity):} The mean absolute deviation between each run's average \textit{LCAS} value and the overall mean score $s_2$. A high $s_3$ indicates that the anomaly signal is highly sensitive to the context, a characteristic often found in unstructured or chaotic artifacts:
    \begin{equation}
        \mu_k = \frac{1}{|M_k|} \sum_{i \in M_k} \ell_i^{(k)}, \qquad
        s_3 = \frac{1}{K} \sum_{k=1}^{K} \left| \mu_k - s_2 \right|.
        \label{eq:s3}
    \end{equation}
\end{itemize}

These statistics are then passed through a learned multi-layer perceptron to produce the image-level anomaly feature vector:
\begin{equation}
f_{\text{anomaly}} = \text{MLP}_{\text{agg}}([s_1, s_2, s_3]) \in \mathbb{R}^{D},
\label{eq:f_anomaly}
\end{equation}
where $\text{MLP}_{\text{agg}}$ is a 3-layer network that projects the low-dimensional statistics into the same $D$-dimensional space as $f_{\text{global}}$ to enable additive fusion.
Collectively, this feature vector $f_{\text{anomaly}}$ provides a rich description of the image's anomalous properties by capturing not just the average magnitude ($s_2$) of the \textit{LCAS} values, but also their distribution ($s_1$) and stability under perturbation ($s_3$). 
\subsection{Feature Fusion and Classification}
\label{sec:fusion}
We fuse the global semantic representation $f_{\text{global}}$ and 
image-level anomaly feature vector $f_{\text{anomaly}}$ through additive 
fusion. This design allows the global representation to be refined by learned anomaly 
corrections, reflecting the intuition that anomalies provide evidence for revising 
the baseline global prediction. After layer normalization, we simply add the anomaly 
features to the global features:
\begin{equation}
f_{\text{corrected}} = f_{\text{global}}^* + f_{\text{anomaly}}^*,
\label{eq:fusion}
\end{equation}
where $(\cdot)^*$ denotes LayerNorm.

This additive strategy makes the correction mechanism explicit rather than forcing the classifier to infer interactions between feature spaces. 
The corrected representation $f_{\text{corrected}}$ is then passed to a classifier for binary prediction; the fusion module and classifier are trained with BCE loss. 
Formally, we compute the scalar logit $z$ using the classifier head $\text{MLP}_{\text{cls}}$:
\begin{equation}
    z = \text{MLP}_{\text{cls}}(f_{\text{corrected}}).
    \label{eq:logit}
\end{equation}

\begin{table}[t]
    \centering
    \setlength{\tabcolsep}{2pt}
    \caption{\textbf{Comparison on GenImage.} Trained on SD v1.4. Accuracies (\%) from AIDE~\cite{yan2025chameleon} and C2P-CLIP~\cite{tan2025c2p}. Best/second-best in \textbf{bold}/\underline{underlined}.}
    \vspace{-4pt}
    \label{tab:genimage_main}
    \scriptsize
    \resizebox{\linewidth}{!}{
    \begin{tabular}{lccccccccc}
        \toprule
        \textbf{Method} &
        \textbf{MJ} & \textbf{SD v1.4} & \textbf{SD v1.5} &
        \textbf{ADM} & \textbf{GLIDE} & \textbf{Wukong} &
        \textbf{VQDM} & \textbf{BigGAN} & \textbf{mAcc} \\
        \midrule
        Swin-T \cite{liu2021swin} & 62.10 & \underline{99.90} & \underline{99.80} & 49.80 & 67.60 & 99.10 & 62.30 & 57.60 & 74.78 \\
        CNNSpot \cite{wang2020cnn} & 52.80 & 96.30 & 95.90 & 50.10 & 39.80 & 78.60 & 53.40 & 46.80 & 64.21 \\
        Spec \cite{zhang2019detecting} & 52.00 & 99.40 & 99.20 & 49.70 & 49.80 & 94.80 & 55.60 & 49.80 & 68.79 \\
        F3Net \cite{qian2020thinking} & 50.10 & \underline{99.90} & \textbf{99.90} & 49.90 & 50.00 & \underline{99.90} & 49.90 & 49.90 & 68.69 \\
        GramNet \cite{liu2020global} & 54.20 & 99.20 & 99.10 & 50.30 & 54.60 & 98.90 & 50.80 & 51.70 & 69.85 \\
        DIRE \cite{wang2023dire} & 60.20 & \underline{99.90} & \underline{99.80} & 50.90 & 55.00 & 99.20 & 50.10 & 50.20 & 70.66 \\
        UnivFD \cite{ojha2023towards} & 73.20 & 84.20 & 84.00 & 55.20 & 76.90 & 75.60 & 56.90 & 80.30 & 73.29 \\
        GenDet \cite{zhu2023gendet} & 89.60 & 96.10 & 96.10 & 58.00 & 78.40 & 92.80 & 66.50 & 75.00 & 81.56 \\
        PatchCraft \cite{zhong2023rich} & 79.00 & 89.50 & 89.30 & 77.30 & 78.40 & 89.30 & 83.70 & 72.40 & 82.30 \\
        AIDE \cite{yan2025chameleon} & 79.38 & 99.74 & 99.76 & 78.54 & 91.82 & 98.65 & 80.26 & 66.89 & 86.88 \\
        LaRE \cite{Luo2024LaRE2} & 74.00 & \textbf{100.00} & \textbf{99.90} & 61.70 & 88.60 & \textbf{100.00} & \underline{97.20} & 67.80 & 86.20 \\
        FatFormer \cite{liu2024fatformer} & \underline{92.70} & \textbf{100.00} & \textbf{99.90} & 75.90 & 88.00 & \underline{99.90} & \textbf{98.80} & 55.80 & 88.90 \\
        C2P-CLIP \cite{tan2025c2p} & 88.20 & 90.90 & 97.90 & \textbf{96.40} & \textbf{99.00} & 98.80 & 96.50 & \textbf{98.70} & \underline{95.80} \\
        \midrule
        \rowcolor{gray!10}
        \textbf{CINEMAE (Ours)} & \textbf{95.90} & 98.78 & 98.50 & \underline{96.07} & \underline{95.93} & 95.96 & 96.01 & \underline{95.90} & \textbf{96.63} \\
        \bottomrule
    \end{tabular}}
    
    \vspace{8pt}
    \caption{\textbf{Comparison on AIGCDetectBenchmark.} Trained on LSUN+ProGAN. Accuracies (\%) from AIDE~\cite{yan2025chameleon}. Best/second-best in \textbf{bold}/\underline{underlined}.}
    \vspace{-4pt}
    \label{tab:cross_gen}
    \resizebox{\linewidth}{!}{
    \begin{tabular}{l*{17}{c}}
        \toprule
        \textbf{Method} &
        \rotatebox{70}{ProGAN} &
        \rotatebox{70}{StyleGAN} &
        \rotatebox{70}{BigGAN} &
        \rotatebox{70}{CycleGAN} &
        \rotatebox{70}{StarGAN} &
        \rotatebox{70}{GauGAN} &
        \rotatebox{70}{StyleGAN2} &
        \rotatebox{70}{WFIR} &
        \rotatebox{70}{ADM} &
        \rotatebox{70}{Glide} &
        \rotatebox{70}{MJ} &
        \rotatebox{70}{SD v1.4} &
        \rotatebox{70}{SD v1.5} &
        \rotatebox{70}{VQDM} &
        \rotatebox{70}{Wukong} &
        \rotatebox{70}{DALLE2} &
        \rotatebox{70}{\textbf{Mean}} \\
        \midrule
        CNNSpot~\cite{wang2020cnn} & \textbf{100.0} & 90.2 & 71.2 & 87.6 & 94.6 & 81.4 & 86.9 & \underline{91.7} & 60.4 & 58.1 & 51.4 & 50.6 & 50.5 & 56.5 & 51.0 & 50.5 & 70.8 \\
        FreDect~\cite{frank2020leveraging} & 99.4 & 78.0 & 82.0 & 78.8 & 94.6 & 80.6 & 66.2 & 50.8 & 63.4 & 54.1 & 45.9 & 38.8 & 39.2 & 77.8 & 40.3 & 34.7 & 64.0 \\
        Fusing~\cite{ju2022fusing} & \textbf{100.0} & 85.2 & 77.4 & 87.0 & 97.0 & 77.0 & 83.3 & 66.8 & 49.0 & 57.2 & 52.2 & 51.0 & 51.4 & 55.1 & 51.7 & 52.8 & 68.4 \\
        LNP~\cite{liu2022detecting} & 99.7 & 91.8 & 77.8 & 84.1 & \underline{99.9} & 75.4 & 94.6 & 70.9 & 84.7 & 80.5 & 65.6 & 85.6 & 85.7 & 74.5 & 82.1 & 88.8 & 83.8 \\
        LGrad~\cite{tan2023lgrad} & 99.8 & 91.1 & 85.6 & 86.9 & 99.3 & 78.5 & 85.3 & 55.7 & 67.2 & 66.1 & 65.4 & 63.0 & 63.7 & 73.0 & 59.6 & 65.5 & 75.3 \\
        UnivFD~\cite{ojha2023towards} & 99.8 & 84.9 & \underline{95.1} & \underline{98.3} & 95.8 & \textbf{99.5} & 75.0 & 86.9 & 66.9 & 62.5 & 56.1 & 63.7 & 63.5 & 85.3 & 70.9 & 50.8 & 78.4 \\
        DIRE-G~\cite{wang2023dire} & 95.2 & 83.0 & 70.1 & 74.2 & 95.5 & 67.8 & 75.3 & 58.1 & 75.8 & 71.8 & 58.0 & 49.7 & 49.8 & 53.7 & 54.5 & 66.5 & 68.7 \\
        PatchCr.~\cite{zhong2023rich} & \textbf{100.0} & 92.8 & \textbf{95.8} & 70.2 & \textbf{100.0} & 71.6 & 89.6 & 85.8 & 82.2 & 83.8 & \underline{90.1} & \textbf{95.4} & \textbf{95.3} & 88.9 & 91.1 & \textbf{96.6} & 89.3 \\
        NPR~\cite{tan2024rethinking} & 99.8 & 97.7 & 84.4 & 96.1 & 99.4 & 82.5 & \textbf{98.4} & 65.8 & 69.7 & 78.4 & 77.9 & 78.6 & 78.9 & 78.1 & 76.1 & 64.9 & 82.9 \\
        AIDE~\cite{yan2025chameleon} & \textbf{100.0} & \textbf{99.6} & 84.0 & \textbf{98.5} & 99.9 & 73.3 & \underline{98.0} & \textbf{94.2} & \textbf{93.4} & \textbf{95.1} & 77.2 & \underline{93.0} & \underline{92.9} & \textbf{95.2} & \textbf{93.6} & \textbf{96.6} & \underline{92.8} \\
        \midrule
        \rowcolor{gray!10}
            \textbf{Ours} & \underline{99.9} & \underline{98.5} & 92.9 & 94.3 & 98.5 & \underline{92.4} & 95.2 & 88.4 & \underline{93.2} & \underline{92.7} & \textbf{91.7} & 92.9 & \underline{92.9} & \underline{92.6} & \underline{92.7} & \underline{94.7} & \textbf{93.96} \\
        \bottomrule
    \end{tabular}}
\end{table}

\section{Experiments}
\label{sec:experiments}
\paragraph{Implementation Details.}
We use a fully frozen MAE ViT-L/16 backbone (both encoder and decoder) pretrained on ImageNet, processing 224$\times$224 images with 16$\times$16 patches and 75\% masking ratio following~\cite{he2022masked}. We set $\lambda=0.1$ and $K=2$ via grid search. Training uses Adam (lr=1e-4), batch size 32, and binary cross-entropy loss for 25 epochs on four NVIDIA RTX 3090 GPUs.
\subsection{Main Results}
We evaluate CINEMAE on two benchmarks: \textbf{GenImage}~\cite{zhu2024genimage} (released March 2024), a million-scale dataset covering 8 generators, and \textbf{AIGCDetectBenchmark}~\cite{zhong2023rich} (April 2025 Ver.), which spans 16 diverse generative models including both GANs and diffusion models.
\paragraph{Cross-Generator Performance.}
On GenImage (trained on SD v1.4), CINEMAE achieves \textbf{96.63\% mAcc}, outperforming all 13 compared methods (\cref{tab:genimage_main}). On AIGCDetectBenchmark (trained on LSUN~\cite{yu2015lsun} + ProGAN~\cite{karras2017progan}), CINEMAE achieves \textbf{93.96\% mAcc} across 16 unseen generators (\cref{tab:cross_gen}).

The key finding is \textbf{generalization consistency}. 
Competitive methods suffer sharp drops on certain generators (\eg, FatFormer~\cite{liu2024fatformer} to 55.80\% on BigGAN, LaRE~\cite{Luo2024LaRE2} to 67.80\% on BigGAN), while CINEMAE maintains \textbf{$\geq$95.9\% on all 8 GenImage generators} and \textbf{$>$91\% on 15 of 16 AIGCDetectBenchmark generators}, indicating that context-based detection captures generator-agnostic signals rather than specific artifacts.

\subsection{Robustness to Real-World Corruptions}
\label{sec:robustness}
We compare against existing methods on AIGCDetectBenchmark under JPEG compression (\cref{tab:robustness_comparison}). 
Most methods suffer catastrophic drops (\eg, LNP: 83.84\% $\rightarrow$ 52.85\% at QF=50). In contrast, \textbf{CINEMAE maintains $>$93\% accuracy across all quality settings}.
This robustness stems from a structural mismatch between JPEG's degradation mechanism and CINEMAE's detection basis: JPEG compression discards high-frequency coefficients through quantization, directly undermining methods that rely on spectral artifacts, whereas CINEMAE's anomaly signal captures patch-level contextual coherence that operates at a semantic level largely preserved under lossy compression.
\begin{table}[t]
\centering
\scriptsize
\caption{\textbf{Robustness to JPEG compression} on AIGCDetectBenchmark (\%). Other results are from AIDE~\cite{yan2025chameleon}.}
\label{tab:robustness_comparison}
\begin{tabular}{l|c|cccc}
    \toprule
    \textbf{Method} & \textbf{Orig.} & \textbf{95} & \textbf{90} & \textbf{75} & \textbf{50} \\
    \midrule
    CNNSpot & 70.78 & 64.03 & 62.26 & 60.65 & 59.66 \\
    FreDect & 64.03 & 66.95 & 67.45 & 66.64 & 65.33 \\
    Fusing & 68.38 & 62.43 & 61.39 & 59.34 & 57.41 \\
    LNP & 83.84 & 53.58 & 54.09 & 53.02 & 52.85 \\
    LGrad & 75.34 & 51.55 & 51.39 & 50.00 & 50.00 \\
    DIRE-G & 68.68 & 66.49 & 66.12 & 65.28 & 64.34 \\
    UnivFD & 78.43 & 74.10 & 74.02 & 69.92 & 68.68 \\
    PatchCraft & 89.31 & 72.48 & 71.41 & 69.43 & 67.78 \\
    AIDE & 92.77 & 75.54 & 74.21 & 70.64 & 69.60 \\
    \midrule
    \rowcolor{gray!10}
    \textbf{CINEMAE} & \textbf{93.96} & \textbf{93.95} & \textbf{93.94} & \textbf{93.92} & \textbf{93.82} \\
    \bottomrule
\end{tabular}
\end{table}

\begin{figure}[b]
\centering
\includegraphics[trim=0 195mm 0 0, clip, width=\linewidth]{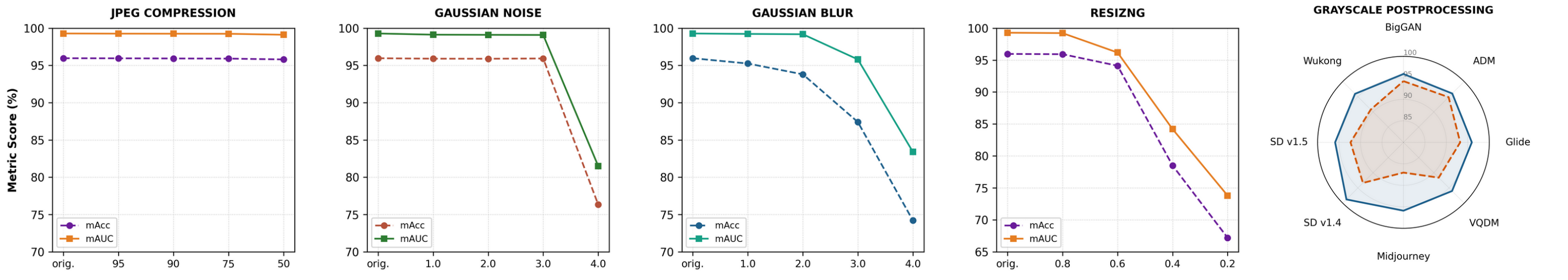}
\caption{Robustness analysis on GenImage. CINEMAE maintains stable mAcc and mAUC under JPEG compression, Gaussian noise, and moderate blur/resize. The radar chart (right) shows per-generator accuracy drop under grayscale conversion.}
\label{fig:robustness}
\end{figure}

\paragraph{Robustness Analysis.}
Additional degradation experiments on GenImage (Fig.~\ref{fig:robustness}) confirm that CINEMAE is invariant to corruptions that preserve semantic structure: JPEG compression causes only 0.14\% drop at QF=50, and Gaussian noise and blur are tolerated at moderate levels. Performance degrades only when corruptions destroy semantic content (\eg, extreme downscaling below scale 0.6). Grayscale conversion causes a notable drop (96.63\% $\rightarrow$ 92.85\% mAcc), revealing that chromatic relationships between patches are themselves a component of the contextual signal.

\subsection{Independence from Low-level Statistics}
\label{sec:correlation}
To further verify that CINEMAE relies on contextual plausibility rather than spectral cues, we compute Spearman correlation coefficients~\cite{spearman1904general} between our anomaly features ($s_1$, $s_2$, $s_3$) and three classical image statistics across the entire GenImage~\cite{zhu2024genimage}: Shannon entropy~\cite{shannon1948mathematical} (pixel intensity complexity), high-frequency power ratio (HF Artifacts~\cite{frank2020leveraging}), and gradient total variation (Grad TV~\cite{rudin1992nonlinear}).

\begin{figure}[h]
    \centering
    \begin{subfigure}{0.42\linewidth}
        \centering
        \caption{Real Images}
        \vspace{2pt}
        \includegraphics[width=\linewidth, trim={0 432pt 0 28pt}, clip]{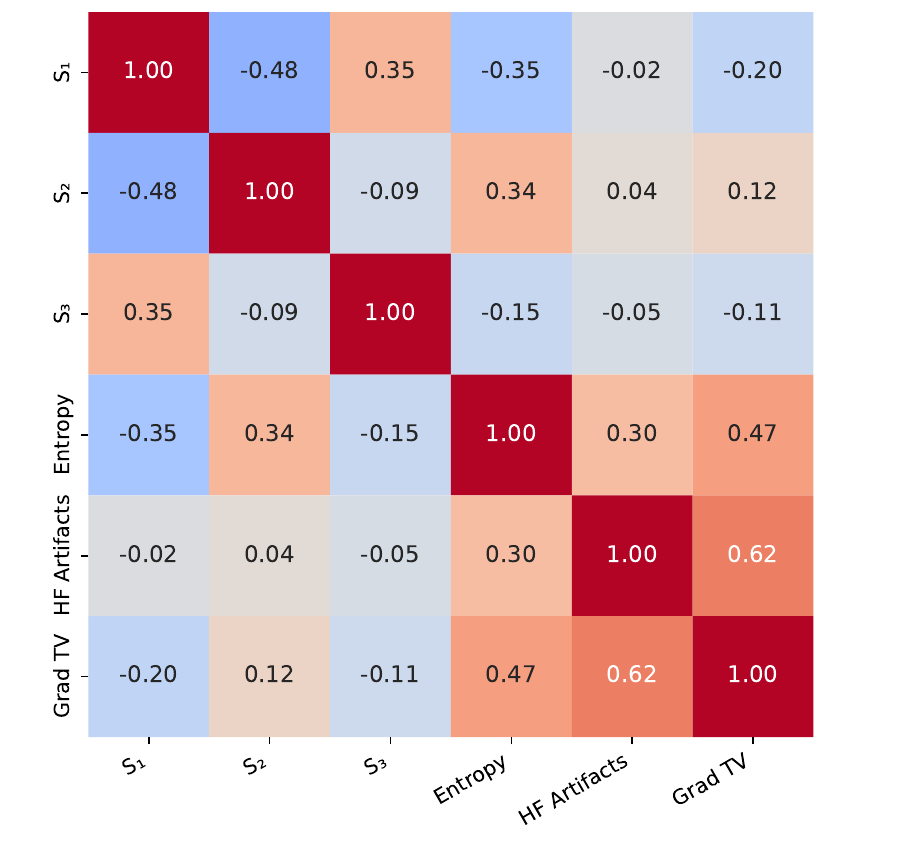}
    \end{subfigure}
    \hspace{0.02\linewidth}
    \begin{subfigure}{0.42\linewidth}
        \centering
        \caption{AI-Generated Images}
        \vspace{2pt}
        \includegraphics[width=\linewidth, trim={0 0 0 460pt}, clip]{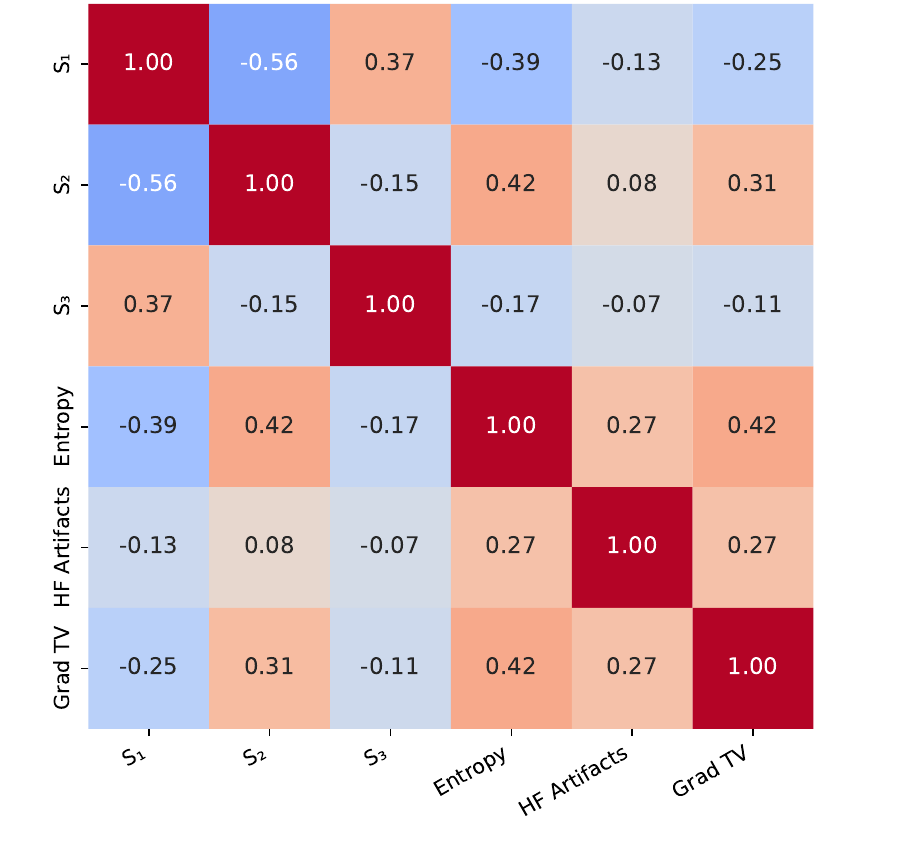}
    \end{subfigure}
    \vspace{-2pt}
    \caption{\textbf{Correlation between CINEMAE features and image statistics.} Negligible correlation with HF Artifacts ($|\rho| < 0.15$) confirms independence from frequency-based detection.}
    \label{fig:correlation_heatmap}
\end{figure}

As shown in Fig.~\ref{fig:correlation_heatmap}, all three components of $f_{\text{anomaly}}$ show negligible correlation with HF Artifacts ($|\rho| < 0.15$), confirming independence from frequency-based detection cues. 
While $s_2$ moderately correlates with global complexity (entropy), $s_3$ remains distinct from all classical metrics, demonstrating that CINEMAE captures unique contextual anomalies invisible to conventional texture or frequency-based analysis. 
This independence directly explains the robustness observed in \cref{sec:robustness}: since CINEMAE does not rely on fragile spectral artifacts, degradations that destroy such cues have minimal impact.

\begin{table*}[b!]
\centering
\caption{\textbf{Evaluation on the Chameleon dataset.}
Overall accuracy and Fake/Real accuracy (\%) on each training split.
CINEMAE achieves balanced fake detection without the strong real-class bias observed in many existing methods.
Best in bold.}
\label{tab:chameleon}
\vspace{-10pt}
\resizebox{\textwidth}{!}{%
\begin{tabular}{l|ccccccccc|c}
\toprule
\textbf{Trained On} & \textbf{CNNSpot} & \textbf{FreDect} & \textbf{Fusing} & \textbf{GramNet} & \textbf{LNP} & \textbf{UnivFD} & \textbf{DIRE} & \textbf{PatchCraft} & \textbf{NPR} & \textbf{CINEMAE} \\
\midrule\midrule
\multirow{2}{*}{SD v1.4} & 60.11 & 56.86 & 57.07 & 60.95 & 55.63 & 55.62 & 59.71 & 56.32 & 58.13 & \multirow{2}{*}{\shortstack{\textbf{61.80} \\ \small{35.50/81.22}}} \\
& \small{8.86/98.63} & \small{1.37/98.57} & \small{0.00/99.96} & \small{17.65/93.50} & \small{0.57/97.01} & \small{74.97/41.09} & \small{11.86/95.67} & \small{3.07/96.35} & \small{2.43/100.00} & \\
\midrule
\multirow{2}{*}{All GenImage} & 60.89 & 57.22 & 57.09 & 59.81 & 58.52 & 60.42 & 57.83 & 55.70 & 57.81 & \multirow{2}{*}{\shortstack{\textbf{62.83} \\ \small{37.10/84.00}}} \\
& \small{9.86/99.25} & \small{0.89/99.55} & \small{0.02/99.98} & \small{8.23/98.58} & \small{7.72/96.70} & \small{85.52/41.56} & \small{2.09/99.73} & \small{1.39/96.52} & \small{1.68/100.00} & \\
\bottomrule
\end{tabular}%
}
\end{table*}

\vspace{-5pt}
\subsection{Qualitative Analysis}
\begin{figure}[!t]
\centering
\includegraphics[width=0.9\linewidth, trim={0 80pt 0 0}, clip]{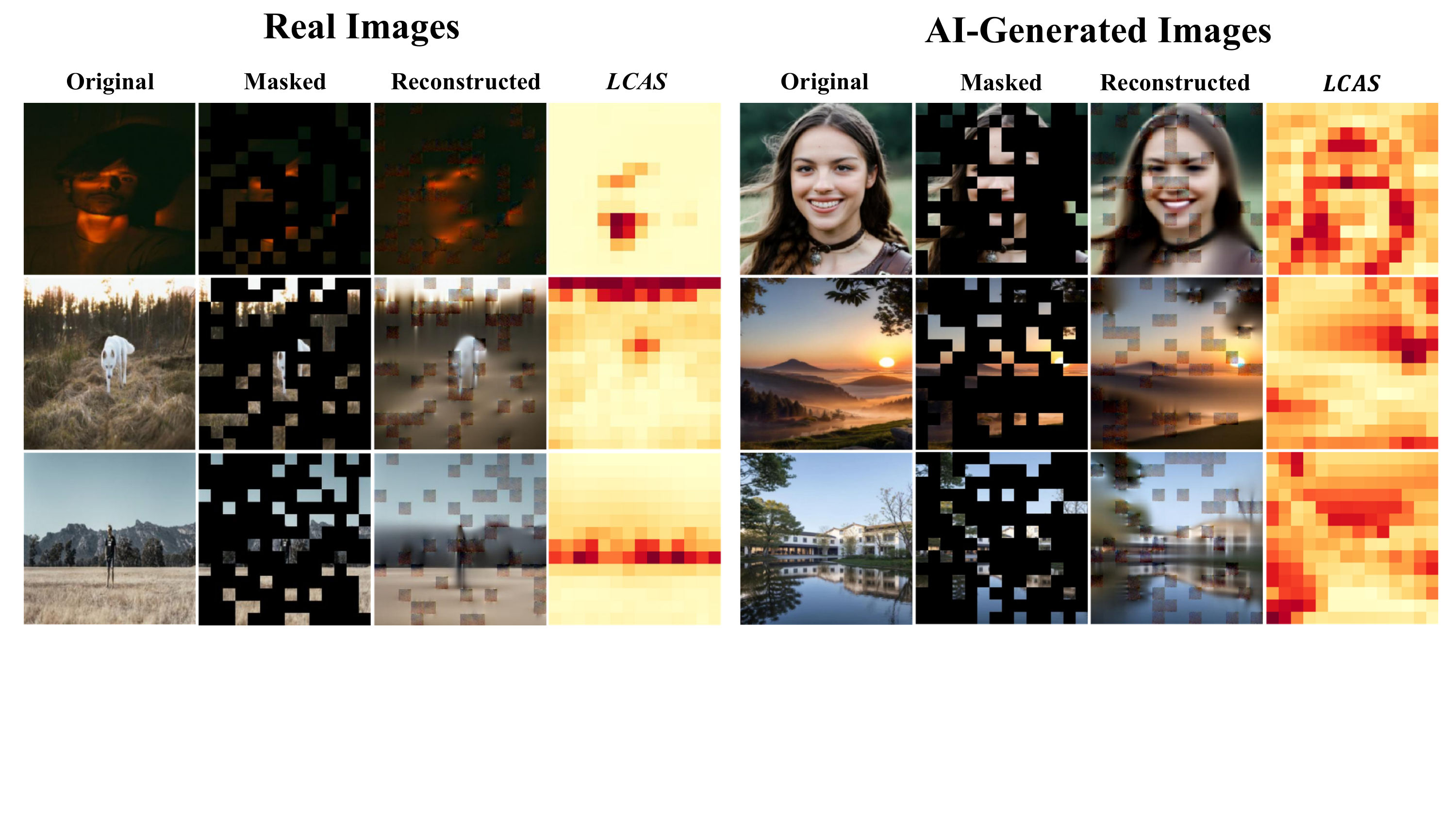}
\vspace{-20pt}
\caption{\textbf{Successful detection results on Chameleon.} \textit{LCAS} values are normalized to [0,1] per image.}
\label{fig:qualitative}
\end{figure}

\paragraph{Challenging Benchmark: Chameleon.} We evaluate CINEMAE on the Chameleon benchmark~\cite{yan2025chameleon}, a curated dataset of perceptually difficult AI-generated images designed to stress-test detector generalization. This benchmark exposes detector biases: most methods exhibit strong real-class bias, achieving high overall accuracy while failing on fakes (\eg, Fusing: 0.02\%; NPR: 1.68\%). In contrast, CINEMAE achieves \textbf{35.50\%} fake detection on the SD v1.4 subset and \textbf{37.10\%} on the full benchmark (\cref{tab:chameleon}), demonstrating more balanced detection.

\cref{fig:qualitative} visualizes successful detection results. In successful cases, real images show high \textit{LCAS} at unpredictable regions (\eg, broken lights, birds in the sky) that are difficult to reconstruct from context, while AI-generated images display uniformly distributed mid-range anomalies from over-predictable pixel statistics.

\begin{figure}[h]
\centering
\includegraphics[width=0.9\linewidth]{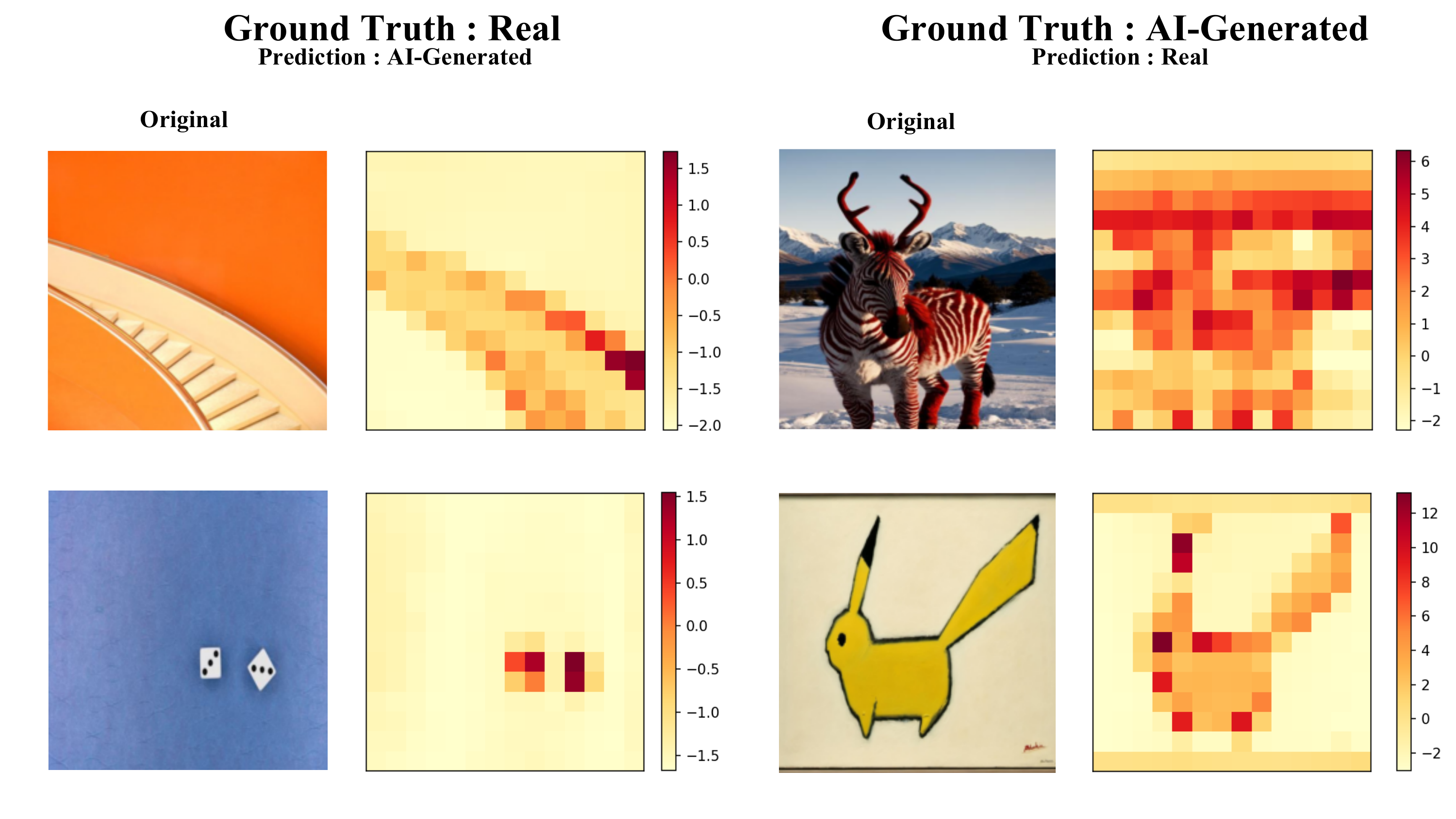}
\vspace{-1pt}
\caption{\textbf{Failure cases.} Left: false positives (simple real images misclassified as AI). Right: false negatives (absurd AI images misclassified as real).}
\label{fig:supp_failure}
\end{figure}

\paragraph{Failure Cases: Low-Context Scenarios.}
\cref{fig:supp_failure} show rare failure modes. \textbf{False positives}: simple images with uniform backgrounds yield low \textit{LCAS} mimicking AI regularity. \textbf{False negatives}: semantically absurd AI images produce high \textit{LCAS}, misinterpreted as natural complexity. These edge cases are exceptional and outside our threat model.

\section{Ablation Studies}
\label{sec:ablation}
In this section, we validate the key design choices of CINEMAE. All models are trained on SD v1.4 and evaluated on GenImage.

\begin{table}[h]
\centering
\caption{\textbf{MAE backbone ablations.} (a) Freezing both encoder and decoder preserves generalization. (b) MAE-L achieves the best trade-off. All in \%.}
\label{tab:ablation_mae}
\scriptsize
\setlength{\tabcolsep}{2pt}
\resizebox{\linewidth}{!}{
\begin{minipage}[t]{0.42\linewidth}
    \centering
    (a) Freeze strategy\\[2pt]
    \begin{tabular}{cc|cc}
        \toprule
        \textbf{Enc.} & \textbf{Dec.} &
        \textbf{mAcc} & \textbf{mAUC} \\
        \midrule
        $\times$     & $\times$     & 71.93 & 51.23  \\
        $\checkmark$ & $\times$     & 73.02 & 53.44  \\
        $\times$     & $\checkmark$ & 74.46 & 68.41 \\
        \rowcolor{gray!10}
        $\checkmark$ & $\checkmark$ & \textbf{96.63} & \textbf{99.28} \\
        \bottomrule
    \end{tabular}
\end{minipage}
\hfill
\begin{minipage}[t]{0.52\linewidth}
    \centering
    (b) MAE backbone\\[2pt]
    \begin{tabular}{l|c|cc}
    \toprule
    \textbf{Model} & \textbf{Params} & \textbf{mAcc} & \textbf{mAUC} \\
    \midrule
    ViT-B/16   & 86M & 73.52 & 81.13    \\
    \rowcolor{gray!10}
    ViT-L/16  & 307M & \textbf{96.63} & \textbf{99.28} \\
    ViT-H/14   & 632M & 88.01 & 86.12 \\
    \bottomrule
    \end{tabular}
\end{minipage}}
\end{table}

\paragraph{MAE freezing strategy.} We ablate MAE encoder/decoder freezing strategies (\cref{tab:ablation_mae}a). Unfreezing either component degrades performance significantly, as the model overfits to training data and loses its natural image prior. Fully freezing both encoder and decoder preserves generalization. 

\paragraph{Reliance on Backbone Capacity.}
We compare MAE-Base, Large, and Huge variants (\cref{tab:ablation_mae}b). MAE-Base underperforms MAE-Large due to limited capacity. However, MAE-Huge also underperforms MAE-Large by 8.62\% mAcc, despite its larger capacity and finer patch size. We attribute this to the capacity-anomaly trade-off: for anomaly detection, the goal is not perfect reconstruction but failing on specific anomalies. MAE-Huge's excessive capacity allows it to reconstruct even high-entropy natural textures, lowering anomaly scores for real images and shrinking the discriminative margin between real and fake distributions.

\begin{table}[h]
\centering
\caption{\textbf{Anomaly signal ablations.} (a) Anomaly statistics components. (b) Sensitivity to $\lambda$: $\lambda=0$ uses $\mathcal{D}_{\text{stat}}$ only (no reconstruction fidelity term). All in \%.}
\label{tab:ablation_anomaly}
\scriptsize
\setlength{\tabcolsep}{3pt}
\begin{minipage}[t]{0.42\linewidth}
    \centering
    (a) Anomaly statistics\\[2pt]
    \begin{tabular}{ccc|cc}
    \toprule
    $s_1$ & $s_2$ & $s_3$ & \textbf{mAcc} & \textbf{mAUC} \\
    \midrule
    $\times$     & $\times$     & $\times$     & 67.12 & 89.71 \\
    $\checkmark$ & $\times$     & $\times$     & 86.83 & 98.64 \\
    $\times$     & $\checkmark$ & $\times$     & 94.12 & 99.08 \\
    $\times$     & $\times$     & $\checkmark$ & 76.95 & 98.07 \\
    $\times$     & $\checkmark$ & $\checkmark$ & 94.61 & 99.10 \\
    $\checkmark$ & $\times$     & $\checkmark$ & 90.77 & 98.96 \\
    $\checkmark$ & $\checkmark$ & $\times$     & 95.48 & 99.15 \\
    \midrule
    \rowcolor{gray!10}
    $\checkmark$ & $\checkmark$ & $\checkmark$ & \textbf{96.63} & \textbf{99.28} \\
    \bottomrule
    \end{tabular}
\end{minipage}%
\hspace{1em}%
\begin{minipage}[t]{0.52\linewidth}
    \centering
    (b) Weighting $\lambda$\\[2pt]
    \begin{tabular}{c|cc}
    \toprule
    $\lambda$ & \textbf{mAcc} & \textbf{mAUC} \\
    \midrule
    0.0 ($\mathcal{D}_{\text{stat}}$ only) & 66.19 & 56.91 \\
    \rowcolor{gray!10}
    0.1 & \textbf{96.63} & \textbf{99.28} \\
    0.3 & 92.10 & 87.10 \\
    0.5 & 88.24 & 86.50 \\
    1.0 & 74.95 & 81.20 \\
    \bottomrule
    \end{tabular}
\end{minipage}
\end{table}

\paragraph{Contribution of Anomaly Statistics.}
We ablate components of $f_{\text{anomaly}}$, training all configurations from scratch (\cref{tab:ablation_anomaly}a). Each component is derived from patch-level \textit{LCAS}: $s_1$ (Variability), $s_2$ (Overall Anomaly), and $s_3$ (Perturbation Sensitivity). $s_2$ is the most critical component, and disabling all (using only $f_{\text{global}}$) severely degrades performance, confirming $f_{\text{anomaly}}$ acts as a booster that amplifies MAE's latent detection capability.
\paragraph{Sensitivity to $\lambda$.}
$\lambda$ balances the two \textit{LCAS} terms: $\mathcal{D}_{\text{stat}}$ (statistical deviation) and reconstruction fidelity (\cref{tab:ablation_anomaly}b). At $\lambda=0$, the model relies exclusively on $\mathcal{D}_{\text{stat}}$, yielding 66.19\% mAcc and near-chance 56.91\% mAUC, indicating that local statistics alone cannot separate real and fake distributions without the semantic grounding provided by reconstruction fidelity. Performance peaks at $\lambda=0.1$ and degrades at higher values, suggesting that the reconstruction term must complement rather than dominate the statistical signal.

\begin{figure*}[t]
\centering
\includegraphics[width=0.9\textwidth]{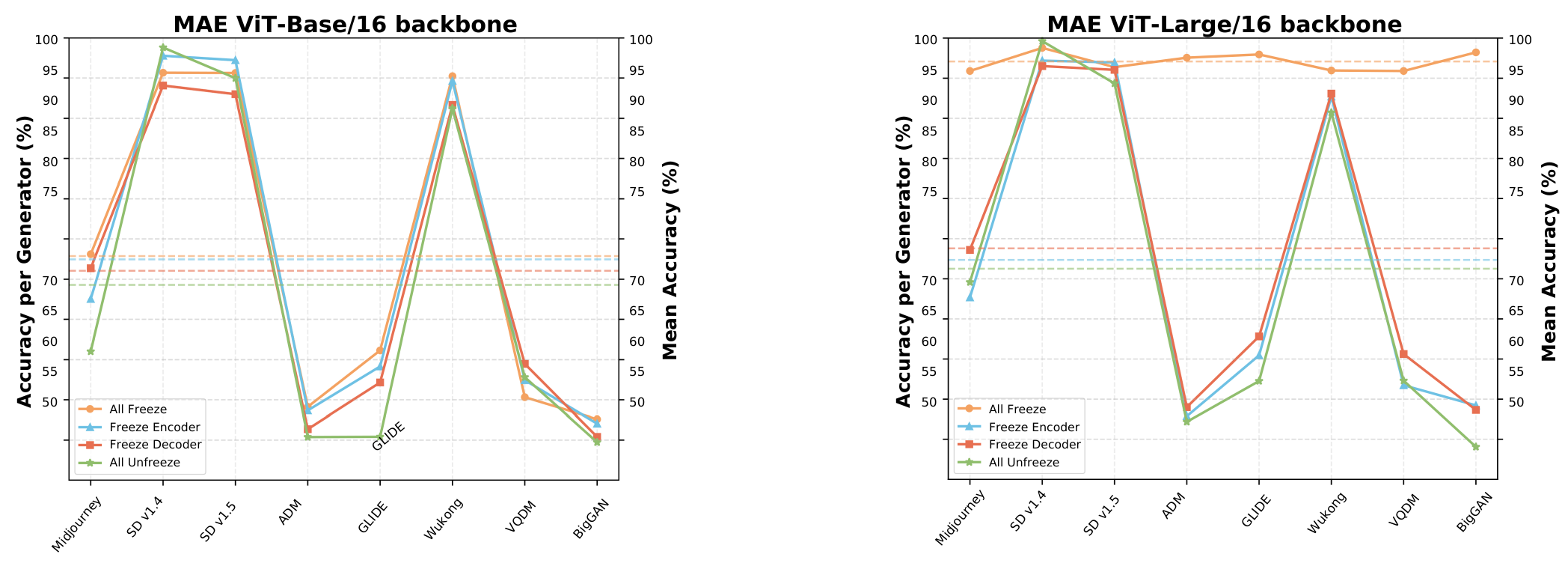}
\caption{\textbf{Ablation on MAE Freezing and Backbone Size.} Trained on SD v1.4, evaluated on GenImage. Solid lines: per-generator accuracy; dotted lines: mean accuracy. Unfreezing any component in ViT-L/16 causes overfitting, while ViT-B/16 underperforms even when fully frozen.}
\label{fig:mae_freezing}
\end{figure*}

\paragraph{Additional Details for MAE Freezing Strategy.}
Both ViT-B/16 and ViT-L/16 suffer discriminative collapse when any component is unfrozen (Fig.~\ref{fig:mae_freezing}), though ViT-B/16 remains limited even when fully frozen due to insufficient expressiveness (Tab.~\ref{tab:ablation_mae}a).

\paragraph{Feature Fusion Strategy.}
We validate our fusion strategy by comparing early and late fusion variants (\cref{tab:ablation_fusion}). Additive fusion clearly outperforms concatenation and gating: allowing $f_{\text{anomaly}}$ to directly refine $f_{\text{global}}$ is more effective than forcing the classifier to learn interactions between feature spaces. Furthermore, early fusion outperforms late fusion (combining features at the decision level), confirming the benefit of feature-level integration.
\begin{table}[h]
    \centering
    \caption{\textbf{Ablation on feature fusion strategy.} Additive fusion treats the anomaly signal as a correction to the global feature. Best in \textbf{bold}.}
    \vspace{-8pt}
    \label{tab:ablation_fusion}
    \scriptsize
    \setlength{\tabcolsep}{2.5pt}
    \begin{tabular}{c|ccc|ccc}
        \toprule
        \textbf{Fusion Level} & \multicolumn{3}{c|}{\textbf{Early Fusion}} & \multicolumn{3}{c}{\textbf{Late Fusion}} \\
        \cmidrule(lr){1-1} \cmidrule(lr){2-4} \cmidrule(lr){5-7}
        \textbf{Method} & Concat & Gate & \textbf{Add (Ours)} & Concat & Gate & Add \\
        \midrule
        mAcc (\%) & 95.54 & 93.95 & \textbf{96.63} & 95.19 & 94.68 & 96.05 \\
        mAUC (\%) & 98.94 & 94.36 & \textbf{99.28} & 98.41 & 97.89 & 99.02 \\
        \bottomrule
    \end{tabular}
\end{table}

\vspace{-2pt}

\section{Conclusion}
We introduced \textbf{CINEMAE}, which adapts the context-based detection paradigm from text to images by leveraging Masked AutoEncoders. CINEMAE formulates MAE's reconstruction error as a contextual anomaly signal and combines it with global semantic features, achieving 96.63\% on GenImage and 93.96\% on AIGCDetectBenchmark while maintaining robustness under JPEG compression. Limitations include degraded performance on low-context images and grayscale inputs, and our formulation approximates rather than exactly models true conditional distributions, though empirically sufficient across diverse generators. Future work includes evaluating robustness under adaptive adversarial attacks and exploring alternative context modeling strategies such as predictive or contrastive formulations. Our work demonstrates that MAE's reconstruction mechanism provides a feasible basis for context-based detection in the image domain.
\bibliographystyle{splncs04}
\bibliography{main}

@String(CVPR  = {CVPR})

@String(ICCV  = {ICCV})

@String(ECCV  = {ECCV})

@String(ICLR  = {ICLR})

@String(ICML  = {ICML})

@String(AAAI  = {AAAI})

@inproceedings{he2022masked,
  author    = {He, Kaiming and Chen, Xinlei and Xie, Saining and Li, Yanghao and Doll{\'a}r, Piotr and Girshick, Ross},
  title     = {Masked Autoencoders Are Scalable Vision Learners},
  booktitle = {Proceedings of the IEEE/CVF Conference on Computer Vision and Pattern Recognition (CVPR)},
  pages     = {16000--16009},
  year      = {2022}
}

@inproceedings{mitchell2023detectgpt,
  title={Detectgpt: Zero-shot machine-generated text detection using probability curvature},
  author={Mitchell, Eric and Lee, Yoonho and Khazatsky, Alexander and Manning, Christopher D and Finn, Chelsea},
  booktitle={International conference on machine learning (ICML)},
  pages={24950--24962},
  year={2023},
  organization={PMLR}
}

@inproceedings{radford2021learning,
  author    = {Radford, Alec and Kim, Jong Wook and Hallacy, Chris and Ramesh, Aditya and Goh, Gabriel and Agarwal, Sandhini and Sastry, Girish and Askell, Amanda and Mishkin, Pamela and Clark, Jack and others},
  title     = {Learning Transferable Visual Models From Natural Language Supervision},
  booktitle = {Proceedings of the 38th International Conference on Machine Learning (ICML)},
  series    = {Proceedings of Machine Learning Research},
  volume    = {139},
  pages     = {8748--8763},
  publisher = {PMLR},
  year      = {2021}
}

@inproceedings{zhu2024genimage,
  author    = {Zhu, Yifan and Chen, Hong and Shen, Weiming and Liu, Zihan and Zhang, Zhen and Liu, Wei},
  title     = {{GenImage}: A Million-Scale Benchmark for Detecting {AI}-Generated Image},
  booktitle = {Advances in Neural Information Processing Systems (NeurIPS)},
  year      = {2023}
}

@inproceedings{du2025forensichub,
  title     = {{ForensicHub}: A Unified Benchmark and Codebase for All-Domain Fake Image Detection and Localization},
  author    = {Du, Bo and Zhu, Xuekang and Ma, Xiaochen and Qu, Chenfan and Feng, Kaiwen and Yang, Zhe and Pun, Chi-Man and Liu, Jian and Zhou, Jizhe},
  booktitle = {Advances in Neural Information Processing Systems (NeurIPS)},
  year      = {2025}
}

@inproceedings{corvi2023detection,
  author    = {Corvi, Riccardo and Cozzolino, Davide and Poggi, Giovanni and Verdoliva, Luisa},
  title     = {On the Detection of Synthetic Images Generated by Diffusion Models},
  booktitle = {ICASSP 2023 - 2023 IEEE International Conference on Acoustics, Speech and Signal Processing (ICASSP)},
  pages     = {1--5},
  year      = {2023}
}

@inproceedings{drct_icml24,
    title={{DRCT}: Diffusion Reconstruction Contrastive Training towards Universal Detection of Diffusion Generated Images},
    author={Baoying Chen and Jishen Zeng and Jianquan Yang and Rui Yang},
    booktitle={Forty-first International Conference on Machine Learning (ICML)},
    year={2024}
}

@inproceedings{ojha2023towards,
  author    = {Ojha, Utkarsh and Li, Yuheng and Lee, Yong Jae},
  title     = {Towards Universal Fake Image Detectors that Generalize Across Generative Models},
  booktitle = {Proceedings of the IEEE/CVF Conference on Computer Vision and Pattern Recognition (CVPR)},
  pages     = {24480--24489},
  year      = {2023}
}

@inproceedings{das2023unmasking,
  title={Unmasking deepfakes: Masked autoencoding spatiotemporal transformers for enhanced video forgery detection},
  author={Das, Sayantan and Kolahdouzi, Mojtaba and {\"O}zparlak, Levent and Hickie, Will and Etemad, Ali},
  booktitle={2023 IEEE International Joint Conference on Biometrics (IJCB)},
  pages={1--11},
  year={2023},
  organization={IEEE}
}

@inproceedings{wang2020cnn,
  author    = {Wang, Sheng-Yu and Wang, Oliver and Zhang, Richard and Owens, Andrew and Efros, Alexei A.},
  title     = {{CNN}-generated Images Are Surprisingly Easy to Spot... for Now},
  booktitle = {Proceedings of the IEEE/CVF Conference on Computer Vision and Pattern Recognition (CVPR)},
  pages     = {8695--8704},
  year      = {2020}
}

@inproceedings{ju2022fusing,
  title={Fusing global and local features for generalized ai-synthesized image detection},
  author={Ju, Yan and Jia, Shan and Ke, Lipeng and Xue, Hongfei and Nagano, Koki and Lyu, Siwei},
  booktitle={2022 IEEE International Conference on Image Processing (ICIP)},
  pages={3465--3469},
  year={2022},
  organization={IEEE}
}

@inproceedings{zed2024eccv,
  title={Zero-shot detection of ai-generated images},
  author={Cozzolino, Davide and Poggi, Giovanni and Nie{\ss}ner, Matthias and Verdoliva, Luisa},
  booktitle={European Conference on Computer Vision (ECCV)},
  pages={54--72},
  year={2024},
  organization={Springer}
}

@inproceedings{yan2025chameleon,
    title={A Sanity Check for {AI}-generated Image Detection},
    author={Shilin Yan and Ouxiang Li and Jiayin Cai and Yanbin Hao and Xiaolong Jiang and Yao Hu and Weidi Xie},
    booktitle={The Thirteenth International Conference on Learning Representations (ICLR)},
    year={2025}
}

@inproceedings{liu2021swin,
  title={Swin transformer: Hierarchical vision transformer using shifted windows},
  author={Liu, Ze and Lin, Yutong and Cao, Yue and Hu, Han and Wei, Yixuan and Zhang, Zheng and Lin, Stephen and Guo, Baining},
  booktitle={Proceedings of the IEEE/CVF international conference on computer vision (ICCV)},
  pages={10012--10022},
  year={2021}
}

@inproceedings{zhang2019detecting,
  title={Detecting and simulating artifacts in gan fake images},
  author={Zhang, Xu and Karaman, Svebor and Chang, Shih-Fu},
  booktitle={2019 IEEE international workshop on information forensics and security (WIFS)},
  pages={1--6},
  year={2019},
  organization={IEEE}
}

@inproceedings{qian2020thinking,
  title={Thinking in frequency: Face forgery detection by mining frequency-aware clues},
  author={Qian, Yuyang and Yin, Guojun and Sheng, Lu and Chen, Zixuan and Shao, Jing},
  booktitle={European conference on computer vision (ECCV)},
  pages={86--103},
  year={2020},
  organization={Springer}
}

@inproceedings{liu2020global,
  title={Global texture enhancement for fake face detection in the wild},
  author={Liu, Yong-Zhen and Li, Guan-Nan and Wu, Yong-Liang and Zhou, Zheng-Jun and Zhao, Yao},
  booktitle={Proceedings of the IEEE/CVF conference on computer vision and pattern recognition (CVPR)},
  pages={8060--8069},
  year={2020}
}

@article{zhu2023gendet,
  author={Mingjian Zhu and Hanting Chen and Mouxiao Huang and Wei Li and Hailin Hu and Jie Hu and Yunhe Wang},
  title={GenDet: Towards Good Generalizations for AI-Generated Image Detection},
  year={2023},
  journal={Computing Research Repository (CoRR)}
}

@article{zhong2023rich,
  title={Rich and poor texture contrast: A simple yet effective approach for ai-generated image detection},
  author={Zhong, Nan and Xu, Yiran and Qian, Zhenxing and Zhang, Xinpeng},
  journal={Computing Research Repository (CoRR)},
  year={2023}
}

@article{Ruderman1994,
  title={The statistics of natural images},
  author={Ruderman, Daniel L},
  journal={Network: computation in neural systems},
  volume={5},
  number={4},
  pages={517},
  year={1994},
  publisher={IOP Publishing}
}

@article{lyu2024deepfake,
  title     = {DeepFake the menace: mitigating the negative impacts of AI-generated content},
  author    = {Lyu, Siwei},
  journal   = {Organizational Cybersecurity Journal: Practice, Process and People},
  volume    = {ahead-of-print},
  year      = {2024},
  publisher = {Emerald Publishing Limited}
}

@InProceedings{Rombach_2022_CVPR,
    author    = {Rombach, Robin and Blattmann, Andreas and Lorenz, Dominik and Esser, Patrick and Ommer, Bj\"orn},
    title     = {High-Resolution Image Synthesis With Latent Diffusion Models},
    booktitle = {Proceedings of the IEEE/CVF Conference on Computer Vision and Pattern Recognition (CVPR)},
    month     = {June},
    year      = {2022},
    pages     = {10684-10695}
}

@inproceedings{wukong2022,
     author = {Gu, Jiaxi and Meng, Xiaojun and Lu, Guansong and Hou, Lu and Minzhe, Niu and Liang, Xiaodan and Yao, Lewei and Huang, Runhui and Zhang, Wei and Jiang, Xin and XU, Chunjing and Xu, Hang},
     booktitle = {Advances in Neural Information Processing Systems (NeurIPS)},
     pages = {26418--26431},
     title = {Wukong: A 100 Million Large-scale Chinese Cross-modal Pre-training Benchmark},
     volume = {35},
     year = {2022}
}

@inproceedings{Luo2024LaRE2,
  title={LaRE\^{} 2: Latent reconstruction error based method for diffusion-generated image detection},
  author={Luo, Yunpeng and Du, Junlong and Yan, Ke and Ding, Shouhong},
  booktitle={Proceedings of the IEEE/CVF Conference on Computer Vision and Pattern Recognition (CVPR)},
  pages={17006--17015},
  year={2024}
}

@InProceedings{Cai_2023_CVPR,
    author    = {Cai, Zhixi and Ghosh, Shreya and Stefanov, Kalin and Dhall, Abhinav and Cai, Jianfei and Rezatofighi, Hamid and Haffari, Reza and Hayat, Munawar},
    title     = {MARLIN: Masked Autoencoder for Facial Video Representation LearnINg},
    booktitle = {Proceedings of the IEEE/CVF Conference on Computer Vision and Pattern Recognition (CVPR)},
    month     = {June},
    year      = {2023},
    pages     = {1493-1504}
}

@InProceedings{Wang_2025_ICASSP,
    author    = {Wang, H. and Liu, Z. and Wang, S.},
    title     = {MIFAE-Forensics: Masked Image-Frequency AutoEncoder for DeepFake Detection},
    booktitle = {ICASSP 2025 - 2025 IEEE International Conference on Acoustics, Speech and Signal Processing (ICASSP)},
    year      = {2025},
    pages     = {1-5},
}

@inproceedings{song2021denoising,
title={Denoising Diffusion Implicit Models},
author={Jiaming Song and Chenlin Meng and Stefano Ermon},
booktitle={International Conference on Learning Representations (ICLR)},
year={2021}
}

@inproceedings{ho2020denoising,
  title={Denoising Diffusion Probabilistic Models},
  author={Ho, Jonathan and Jain, Ajay and Abbeel, Pieter},
  booktitle={Advances in Neural Information Processing Systems (NeurIPS)},
  volume={33},
  pages={6840--6851},
  year={2020}
}

@inproceedings{brock2018large,
  title={Large Scale GAN Training for High Fidelity Natural Image Synthesis},
  author={Brock, Andrew and Donahue, Jeff and Simonyan, Karen},
  booktitle={International Conference on Learning Representations (ICLR)},
  year={2019}
}

@inproceedings{frank2020leveraging,
  title={Leveraging Frequency Analysis for Deep Fake Image Recognition},
  author={Frank, Joel and Eisenhofer, Thorsten and Schönherr, Lea and Fischer, Asja and Kolossa, Dorothea and Holz, Thorsten},
  booktitle={Proceedings of the 37th International Conference on Machine Learning (ICML)},
  pages={3247--3258},
  year={2020},
  series={Proceedings of Machine Learning Research},
  volume={119},
  publisher={PMLR}
}

@inproceedings{liu2022detecting,
  title={Detecting generated images by real images},
  author={Liu, Bo and Yang, Fan and Bi, Xiuli and Xiao, Bin and Li, Weisheng and Gao, Xinbo},
  booktitle={European Conference on Computer Vision (ECCV)},
  pages={95--110},
  year={2022},
  organization={Springer}
}

@inproceedings{wang2023dire,
  title={Dire for diffusion-generated image detection},
  author={Wang, Zhendong and Bao, Jianmin and Zhou, Wengang and Wang, Weilun and Hu, Hezhen and Chen, Hong and Li, Houqiang},
  booktitle={Proceedings of the IEEE/CVF International Conference on Computer Vision (ICCV)},
  pages={22445--22455},
  year={2023}
}

@inproceedings{tan2024rethinking,
  title={Rethinking the up-sampling operations in cnn-based generative network for generalizable deepfake detection},
  author={Tan, Chuangchuang and Zhao, Yao and Wei, Shikui and Gu, Guanghua and Liu, Ping and Wei, Yunchao},
  booktitle={Proceedings of the IEEE/CVF Conference on Computer Vision and Pattern Recognition (CVPR)},
  pages={28130--28139},
  year={2024}
}

@inproceedings{li2025safe,
  title={Improving synthetic image detection towards generalization: An image transformation perspective},
  author={Li, Ouxiang and Cai, Jiayin and Hao, Yanbin and Jiang, Xiaolong and Hu, Yao and Feng, Fuli},
  booktitle={Proceedings of the 31st ACM SIGKDD Conference on Knowledge Discovery and Data Mining V. 1},
  pages={2405--2414},
  year={2025}
}

@inproceedings{liu2024fatformer,
  author = {Liu, Huan and Tan, Zichang and Tan, Chuangchuang and Wei, Yunchao and Wang, Jingdong and Zhao, Yao},
  title = {Forgery-aware Adaptive Transformer for Generalizable Synthetic Image Detection},
  booktitle = {Proceedings of the IEEE/CVF Conference on Computer Vision and Pattern Recognition (CVPR)},
  year = {2024},
  pages = {10770--10780},
  month = {June}
}

@inproceedings{tan2025c2p,
  author = {Tan, Chuangchuang and Tao, Renshuai and Liu, Huan and Gu, Guanghua and Wu, Baoyuan and Zhao, Yao and Wei, Yunchao},
  title = {C2P-CLIP: Injecting Category Common Prompt in CLIP to Enhance Generalization in Deepfake Detection},
  booktitle = {Proceedings of the AAAI Conference on Artificial Intelligence},
  year = {2025},
  pages = {7184--7192}
}

@inproceedings{deng2009imagenet,
title={ImageNet: A large-scale hierarchical image database},
author={Deng, Jia and Dong, Wei and Socher, Richard and Li, Li-Jia and Li, Kai and Fei-Fei, Li},
booktitle={2009 IEEE Conference on Computer Vision and Pattern Recognition (CVPR)},
pages={248--255},
year={2009},
organization={IEEE}
}

@inproceedings{karras2017progan,
  title={Progressive growing of gans for improved quality, stability, and variation},
  author={Karras, Tero and Aila, Timo and Laine, Samuli and Lehtinen, Jaakko},
  booktitle={International Conference on Learning Representations (ICLR)},
  year={2018}
}

@article{yu2015lsun,
  title={Lsun: Construction of a large-scale image dataset using deep learning with humans in the loop},
  author={Yu, Fisher and Seff, Ari and Zhang, Yinda and Song, Shuran and Funkhouser, Thomas and Xiao, Jianxiong},
  journal={arXiv preprint arXiv:1506.03365},
  year={2015}
}

@inproceedings{rossler2019faceforensics++,
  title={Faceforensics++: Learning to detect manipulated facial images},
  author={Rossler, Andreas and Cozzolino, Davide and Verdoliva, Luisa and Riess, Christian and Thies, Justus and Nie{\ss}ner, Matthias},
  booktitle={Proceedings of the IEEE/CVF international conference on computer vision (ICCV)},
  pages={1--11},
  year={2019}
}

@inproceedings{afchar2018mesonet,
  title={Mesonet: a compact facial video forgery detection network},
  author={Afchar, Darius and Nozick, Vincent and Yamagishi, Junichi and Echizen, Isao},
  booktitle={2018 IEEE international workshop on information forensics and security (WIFS)},
  pages={1--7},
  year={2018},
  organization={IEEE}
}

@inproceedings{zhou2018learning,
  title={Learning rich features for image manipulation detection},
  author={Zhou, Peng and Han, Xintong and Morariu, Vlad I and Davis, Larry S},
  booktitle={Proceedings of the IEEE conference on computer vision and pattern recognition (CVPR)},
  pages={1053--1061},
  year={2018}
}

@inproceedings{kwon2021cat,
  title={CAT-Net: Compression artifact tracing network for detection and localization of image splicing},
  author={Kwon, Myung-Joon and Yu, In-Jae and Nam, Seung-Hun and Lee, Heung-Kyu},
  booktitle={Proceedings of the IEEE/CVF winter conference on applications of computer vision (WACV)},
  pages={375--384},
  year={2021}
}

@inproceedings{tan2023lgrad,
  title={Learning on gradients: Generalized artifacts representation for gan-generated images detection},
  author={Tan, Chuangchuang and Zhao, Yao and Wei, Shikui and Gu, Guanghua and Wei, Yunchao},
  booktitle={Proceedings of the IEEE/CVF Conference on Computer Vision and Pattern Recognition (CVPR)},
  pages={12105--12114},
  year={2023}
}

@article{shannon1948mathematical,
  title={A mathematical theory of communication},
  author={Shannon, Claude E},
  journal={Bell System Technical Journal},
  volume={27},
  number={3},
  pages={379--423},
  year={1948}
}

@article{rudin1992nonlinear,
  title={Nonlinear total variation based noise removal algorithms},
  author={Rudin, Leonid I and Osher, Stanley and Fatemi, Emad},
  journal={Physica D: Nonlinear Phenomena},
  volume={60},
  number={1-4},
  pages={259--268},
  year={1992}
}

@article{spearman1904general,
  title={The proof and measurement of association between two things},
  author={Spearman, Charles},
  journal={The American Journal of Psychology},
  volume={15},
  number={1},
  pages={72--101},
  year={1904}
}

@misc{midjourney2022,
  author = {Midjourney},
  title = {Midjourney},
  year = {2023},
  howpublished = {\url{https://www.midjourney.com/}},
  note = {Accessed: 2025-11-21}
}
\end{document}